\pdfoutput=1
\documentclass[letterpaper, 10 pt, conference]{ieeeconf}  

\IEEEoverridecommandlockouts                              

\usepackage{amsmath}
\usepackage{amssymb}
\usepackage{mathtools}
\usepackage{bm}
\usepackage[binary-units]{siunitx}

\usepackage{subcaption}
\usepackage{cleveref}
\usepackage{graphicx}
\usepackage{graphics}
\usepackage{caption}

\usepackage[utf8]{inputenc}
\usepackage{pgfplots}
\DeclareUnicodeCharacter{2212}{−}
\usepgfplotslibrary{groupplots,dateplot}
\usetikzlibrary{patterns,shapes.arrows}
\usepgflibrary{patterns}
\pgfplotsset{compat=newest}
\usepackage{filecontents}
\usepgfplotslibrary{statistics}

\usepackage{pdfpages}

\usepackage{layouts}
\usepackage{makecell}

\usepackage{booktabs}
\usepackage{multirow}
\usepackage{bigstrut}

\usepackage{algorithm2e}

\usepackage{float}

\usepackage{xargs}                      
\usepackage{xcolor}  
\usepackage[colorinlistoftodos,prependcaption,textsize=tiny,obeyDraft]{todonotes}
\usepackage{etoolbox}

\newcommandx{\unsure}[2][1=]{\marginparwidth0.5in\todo[linecolor=red,backgroundcolor=red!25,bordercolor=red#1]{#2}}
\newcommandx{\change}[2][1=]{\marginparwidth0.5in\todo[linecolor=blue,backgroundcolor=blue!25,bordercolor=blue,#1]{#2}}
\newcommandx{\info}[2][1=]{\marginparwidth0.5in\todo[linecolor=OliveGreen,backgroundcolor=OliveGreen!25,bordercolor=OliveGreen,#1]{#2}}
\newcommandx{\improvement}[2][1=]{\marginparwidth0.5in\todo[linecolor=lime,backgroundcolor=lime!25,bordercolor=lime,#1]{#2}}
\newcommandx{\thiswillnotshow}[2][1=]{\marginparwidth0.5in\todo[disable,#1]{#2}}

\newtoggle{arxiv}
\toggletrue{arxiv}

\newcommand{\Tp}{	 		
	^\intercal
}

\newcommand{\Map}[2]{		
	{#1}\mapsto{#2}
}

\newcommand{\Path}[1]{
    \prescript{}{\mathrm{R}}{#1}
}
\newcommand{\Norm}[1]{
    \prescript{}{\mathrm{N}}{#1}
}
\newcommand{\Inp}{								
	z
}

\newcommand{\Time}{								
	t
}
\newcommand{\TimeNorm}{							
	\tau
}

\newcommand{\Vect}[1]{							
	\bm{\mathrm{#1}}
}
\newcommand{\RowVect}[1]{						
	\left[\begin{matrix}#1\end{matrix}\right]
}
\newcommand{\UnitVect}[1]{						
	\Vect{e}_{#1}
}

\newcommand{\Real}{ 	
	\mathbb{R}
}
\newcommand{\RealPos}{	
	\mathbb{R}^+
}
\newcommand{\Nat}{		
	\mathbb{N}
}
\newcommand{\NatPos}{	
	\Nat_0
}

\newcommand{\OrderBasis}{		
	o
}
\newcommand{\Nknots}{		
	r
}
\newcommand{\Brk}{			
	\rho
}
\newcommand{\Brks}{			
	\Vect{\Brk}
}
\newcommand{\Knot}{ 		
	k
}
\newcommand{\Knots}{		
	\Vect{\Knot}
}
\newcommand{\KnotIval}{		
	\Delta\Knot
}
\newcommand{\KnotIvals}{	
	\Delta\Vect{\Knot}
}
\newcommand{\Cont}{			
	v
}
\newcommand{\Conts}{		
	\Vect{\Cont}
}
\newcommand{\Basis}{		
	b
}

\newcommand{\Gab}{			
	\xi
}

\newcommand{\DimBasisSet}{	
	\delta
}
\newcommand{\KnotIdxSet}{	
	\mathcal{I}
}

\newcommand{\Spline}{		
	s
}

\newcommand{\Coef}{			
	\sigma
}
\newcommand{\Coefs}{		
	\Vect{\Coef}
}
\newcommand{\TfCoef}[2]{	
	\mathbf{T}^{#1}_{#2}
}

\newcommand{\CtrlPts}{
	c
}
\newcommand{\CtrlPtsIdxSet}{  
	\Pi
}

\newcommand{\DirLon}{   
    x
}
\newcommand{\DirLat}{   
    y
}

\newcommand{\Pos}{      
    p
}

\newcommand{\PosPath}{  
    \Path{\Pos}
}

\newcommand{\PosPathX}{ 
    \PosPath_\DirLon
}

\newcommand{\PosPathY}{ 
    \PosPath_\DirLat
}

\newcommand{\Vel}{      
    v
}
\newcommand{\VelX}{ 
    \Vel_\DirLon
}

\newcommand{\VelPath}{  
    \Path{\Vel}
}
\newcommand{\VelPathX}{ 
    \VelPath_\DirLon
}
\newcommand{\VelPathY}{ 
    \VelPath_\DirLat
}

\newcommand{\Acc}{      
    a
}

\newcommand{\Jerk}{      
    j
}

\newcommand{\StateVect}{    
    \Vect{x}
}
\newcommand{\StateVectX}{   
    \StateVect_\DirLon
}
\newcommand{\StateVectY}{   
    \StateVect_\DirLat
}

\newcommand{\LenTjy}{       
    T
}
\newcommand{\SplineState}{  
    \Vect{\Spline}
}

\newcommand{\SplinePos}[1]{       
    \Spline_{\Pos_{#1}}
}

\newcommand{\SplinePosVar}[1]{  
    \Spline_{\tilde{\Pos}_{#1}}
}

\newcommand{\SplineVel}[1]{       
    \Spline_{\Vel_{#1}}
}
\newcommand{\SplineVelX}{       
    \Spline_{\Vel_\DirLon}
}
\newcommand{\SplineVelY}{       
    \Spline_{\Vel_\DirLat}
}
\newcommand{\SplineAcc}[1]{       
    \Spline_{\Acc_{#1}}
}
\newcommand{\SplineAccX}{       
    \Spline_{\Acc_\DirLon}
}
\newcommand{\SplineAccY}{       
    \Spline_{\Acc_\DirLat}
}
\newcommand{\SplineJerk}[1]{       
    \Spline_{\Jerk_{#1}}
}

\newcommand{\HeadBound}{    
    \overline{\Head}
}
\newcommand{\HeadBndFct}{  
    \eta
}
\newcommand{\VelBound}{       
    \overline{\Vel}
}
\newcommand{\AccBound}{       
    \overline{\Acc}
}

\newcommand{\CurvPathMax}{      
    \hat{\Curv}_\mathrm{R}
}
\newcommand{\PosYMax}{          
    \hat{\Pos}_y
}

\newcommand{\Head}{  
    \psi
}
\newcommand{\HeadPath}{  
    \Path{\Head}
}

\newcommand{\Curv}{     
    \kappa
}
\newcommand{\CurvPath}{ 
    \Curv_\mathrm{R}
}

\newcommand{\ConstrBasis}{
    \Vect{f}
}

\newcommand{\ConstrBasisHead}[1]{
    \ConstrBasis_{\HeadBound_{#1}}
}
\newcommand{\ConstrBasisHeadLb}{    
    \ConstrBasisHead{\mathrm{lb}}
}
\newcommand{\ConstrBasisHeadUb}{    
    \ConstrBasisHead{\mathrm{ub}}
}

\newcommand{\NumOv}{    
    n_\mathrm{ov}
}
\newcommand{\AxisLen}[1]{
    \Delta_{#1}
}
\newcommand{\AxisLenOv}[1]{
    \Delta_{#1,m}
}

\newcommand{\AxisLenX}{ 
    \AxisLen{\DirLon}
}

\newcommand{\AxisLenXov}{ 
    \AxisLenOv{\DirLon}
}

\newcommand{\AxisLenY}{ 
    \AxisLen{\DirLat}
}

\newcommand{\AxisLenYov}{ 
    \AxisLenOv{\DirLat}
}

\newcommand{\SplinePh}{  
    \Spline_{\square}
}
\newcommand{\GabPh}{  
    \Gab_{\square}
}
\newcommand{\SplineGabPh}{ 
    \prescript{\Gab}{}{\Spline_{\square}}
}

\newcommand{\PosOvVar}[1]{ 
    \tilde{\Pos}_{{#1},m}
}
\newcommand{\PosPathOv}[1]{ 
    \PosPath_{#1,m}
}

\newcommand{\PosPathXov}{   
    \PosPath_{\DirLon,m}
}

\newcommand{\PosPathYov}{   
    \PosPath_{\DirLat,m}
}

\newcommand{\EllipseOv}{  
    e_m
}
\newcommand{\SplinePosOv}[1]{ 
    \SplinePos{#1,m}
}
\newcommand{\SplinePosOvVar}[1]{
    \SplinePosVar{#1,m}
}
\newcommand{\SplineGabPosOvVar}[1]{
    \prescript{\Gab}{}{\SplinePosVar{#1,m}}
}
\newcommand{\SplinePosXov}{ 
    \Spline_{\Pos_{\DirLon,m}}
}
\newcommand{\SplinePosYov}{ 
    \Spline_{\Pos_{\DirLat,m}}
}
\newcommand{\PosDiffOv}[1]{    
    d_{#1,m}
}
\newcommand{\PosDiffSqOv}[1]{  
    d^2_{#1,m}
}
\newcommand{\SplinePosDiffOv}[1]{  
    \Spline_{\PosDiffOv{#1}}
}
\newcommand{\SplineGabPosDiffOv}[1]{  
    \prescript{\Gab}{}{\Spline_{\PosDiffOv{#1}}}
}
\newcommand{\SplinePosDiffSqOv}[1]{  
    \Spline_{\PosDiffSqOv{#1}}
}
\newcommand{\SplineGabPosDiffSqOv}[1]{  
    \prescript{\Gab}{}{\Spline_{\PosDiffSqOv{#1}}}
}
\newcommand{\SplineEllipseOv}{  
    \Spline_{\EllipseOv}
}
\newcommand{\SplineEllipseOvQ}{  
    \tilde{\Spline}_{\EllipseOv}
}
\newcommand{\SplineGabEllipseOvQ}{  
    \prescript{\Gab}{}{\SplineEllipseOvQ}
}

\newcommand{\GabsPosOvVar}[1]{
    \Gab_{\PosOvVar{#1}}
}
\newcommand{\GabPosDiffOv}[1]{ 
    \Gab_{\PosDiffOv{#1}}
}
\newcommand{\GabPosDiffSqOv}[1]{ 
    \Gab_{\PosDiffSqOv{#1}}
}
\newcommand{\GabEllipseOv}{  
    \Gab_{\EllipseOv}
}

\newcommand{\ConstrBasisPosOvVar}[1]{
    \ConstrBasis_{\PosOvVar{#1}}
}
\newcommand{\ConstrBasisPosDiffOv}[1]{
    \ConstrBasis_{\PosDiffOv{#1}}
}
\newcommand{\ConstrBasisPosDiffSqOv}[1]{
    \ConstrBasis_{\PosDiffSqOv{#1}}
}
\newcommand{\ConstrBasisEllipseOv}{
    \ConstrBasis_{\EllipseOv}
}

\newcommand{\GabHeadBound}[1]{ 
    \Gab_{\HeadBound_{#1}}
}
\newcommand{\GabHeadUb}{  
    \GabHeadBound{\mathrm{ub}}
}
\newcommand{\GabHeadLb}{  
    \GabHeadBound{\mathrm{lb}}
}

\newcommand{\SplineHeadBound}[1]{ 
    \Spline_{\HeadBound_{#1}}
}
\newcommand{\SplineHeadLb}{ 
    \SplineHeadBound{\mathrm{lb}}
}\newcommand{\SplineGabHeadLb}{ 
    \prescript{\Gab}{}{\SplineHeadLb}
}
\newcommand{\SplineHeadUb}{ 
    \SplineHeadBound{\mathrm{ub}}
}
\newcommand{\SplineGabHeadUb}{ 
    \prescript{\Gab}{}{\SplineGabHeadUb}
}

\newcommand{\JerkSq}[1]{   
    \Jerk^2_{#1}
}
\newcommand{\SplineJerkSq}[1]{ 
    \Spline_{\JerkSq{#1}}
}

\newcommand{\ConstrBasisJerkSq}[1]{  
    \ConstrBasis_{\JerkSq{#1}}
}

\newcommand{\VelTarget}{  
    \tilde{\Vel}
}
\newcommand{\ConstrTerm}[2]{  
    \Vect{f}_{\mathrm{T}#1,#2}
}
\newcommand{\KnotIdxSetTerm}[2]{    
    \KnotIdxSet_{\mathrm{T}#1,#2}
}

\newcommand{\CoefsOpt}{ 
    \Sigma
}

\newcommand{\KnotIvalMin}{ 
    \Delta\TimeNorm_\mathrm{min}
}

\title{\LARGE \bf
Time-Optimal Trajectory Planning in Highway Scenarios using Basis-Spline Parameterization
}

\author{Philip Dorpmüller$^{1}$, Thomas Schmitz$^{2}$, Naveen Bejagam$^{2}$ and Torsten Bertram$^{1}$
\thanks{$^{1}$TU Dortmund University, Institute of Control Theory and Systems Engineering, 44227 Dortmund, Germany
	{\tt\small philip.dorpmueller@tu-dortmund.de}}%
\thanks{$^{2}$ZF Group, Automated Driving and Integral Cognitive Safety, 40547 Düsseldorf, Germany}%
}

\begin{document}
\bstctlcite{IEEEexample:BSTcontrol}
\iftoggle{arxiv}{
\begin{minipage}{0.95\textwidth}\ \\[12pt]
    \begin{center}
         This paper has been accepted for publication in \textit{2023 IEEE 26th International Conference on Intelligent Transportation Systems (ITSC)}.
    \end{center}
    \vspace{1in}
    ©2023 IEEE. Personal use of this material is permitted. Permission from IEEE must be obtained for all other uses, in any current or future media, including reprinting/republishing this material for advertising or promotional purposes, creating new collective works, for resale or redistribution to servers or lists, or reuse of any copyrighted component of this work in other works.
\end{minipage}
\newpage
}

\maketitle
\thispagestyle{empty}
\pagestyle{empty}

\begin{abstract}
    Basis splines enable a time-continuous feasibility check with a finite number of constraints.
    Constraints apply to the whole trajectory for motion planning applications that require a collision-free and dynamically feasible trajectory.
    Existing motion planners that rely on gradient-based optimization apply time scaling to implement a shrinking planning horizon.
    They neither guarantee a recursively feasible trajectory nor enable reaching two terminal manifold parts at different time scales.
    This paper proposes a nonlinear optimization problem that addresses the drawbacks of existing approaches.
    Therefore, the spline breakpoints are included in the optimization variables.
    Transformations between spline bases are implemented so a sparse problem formulation is achieved.
    A strategy for breakpoint removal enables the convergence into a terminal manifold.
    The evaluation in an overtaking scenario shows the influence of the breakpoint number on the solution quality and the time required for optimization.
\end{abstract}


\section{Introduction}
\label{sec:introduction}
\graphicspath{{figures/introduction/}}

Automated driving is often approached by a modular system architecture \cite{padenSurveyMotionPlanning2016}.
Therein, the motion planner generates a collision-free vehicle trajectory to accomplish local navigation goals defined by the behavior layer.
Furthermore, the planned trajectory must be dynamically feasible and comfortable for the passenger.
Trajectory planning can be viewed as optimizing the function space of trajectories subject to boundary and inequality constraints \cite{padenSurveyMotionPlanning2016}.
The Direct Method is applied here to transcribe the variational problem, which is solved by a gradient-based optimization algorithm.
Compared to the Indirect Method and Dynamic Programming, it yields a high-quality solution considering general inequality constraints in a high dimensional state space \cite{diehlFastDirectMultiple2006}.
This work focuses on the application to highway scenarios.
It is assumed that combinatorial decisions have already been made in the behavior layer.
Thus, a target velocity and lane are passed to the motion planner.
Also, a suboptimal and possibly infeasible initial guess for the optimal trajectory is provided.

Constraints related to dynamic feasibility and collision apply over the whole trajectory rendering the parameter optimization problem semi-infinite \cite{lengagneGenerationDynamicMotions2010}.
Often, it is turned into a finite-dimensional one by applying the constraints to a finite number of time steps.\improvement{removed not required text}
In this case, the trajectory's feasibility cannot be guaranteed in a continuous-time sense.
Alternatively, one can use a basis spline (B-spline) representation of the trajectory to \cite{vanloockBsplineParameterizedOptimal2015}.
A B-spline is constructed from a set of polynomial basis functions defined over a knot vector \cite{boorPracticalGuideSplines2001}.
The knots coincide with the spline's breakpoint (BP) values, duplicated according to the respective BP order of continuity.
The weighted sum over the basis functions yields the B-spline contained in the convex hull of the basis weights.
This property enables the formulation of a finite-dimensional optimization problem without losing the continuous time feasibility.

The planning problem can be formulated in a receding or shrinking horizon manner \cite{diehlFastDirectMultiple2006}.
\cite{kunzHybridDiscreteparametricOptimization2016} uses the second approach to plan time-consistent polynomial trajectories in highway scenarios into a terminal manifold.
In addition, the target velocity and target lane can be reached at two different times.
The shrinking horizon problem introduced by \cite{vanloockBsplineParameterizedOptimal2015} uses time scaling of the B-spline BPs.
This approach fails to ensure the recursive feasibility of the solution trajectory.
The property is desirable since it guarantees a feasible solution in the next step if one is found in the current one \cite{gruneNonlinearModelPredictive2011}.
Compliance of the resulting trajectories with Bellman's principle of optimality is one option to enable the property.
\cite{rosmannStabilisingQuasitimeoptimalNonlinear2022} discusses recursive feasibility for the time discrete shrinking horizon.
Since all BPs are scaled uniformly in \cite{vanloockBsplineParameterizedOptimal2015}, the time intervals in between decrease in closed-loop.
Contrary, only a refinement of the existing BPs is allowed to yield consistent planning results \cite{boorPracticalGuideSplines2001}.
Thus, the solution trajectory in the next step might not be consistent with the current solution, and Bellman's principle of optimality does not hold.
Also, a single time scale cannot realize different time horizons for the longitudinal and lateral directions.

This paper contributes a nonlinear optimization problem to plan optimal trajectories in the context of highway driving into a selected terminal manifold.\improvement{replace time-optimal with terminal manifold}
The problem adopts a B-spline parameterization and guarantees recursive feasibility.
The property is achieved with a novel BP removal strategy and the optimization of the spline BPs.
The time into the terminal manifold is used as an optimization criterion.
The optimization algorithm performs coefficient transformations implicitly to retain a sparse problem formulation.\improvement{complete contribution}
Furthermore, two different time scales are introduced for the longitudinal and lateral directions.

The following section \ref{sec:related_work} provides an overview of the literature dealing with spline-based trajectory planning in highway environments and applying B-splines in motion planning.
Afterward, a background on B-splines is given in section \ref{sec:basis_splines}.
Section \ref{sec:problem_formulation} develops the nonlinear optimization problem.
A strategy to successively remove BPs during the simulation is described in section \ref{sec:breakpoint_adaption}.
The proposed approach is evaluated in section \ref{sec:evaluation} in a highway scenario using different numbers of BPs.
The final section \ref{sec:summary} summarizes the results and discusses future work.
\section{Related Work}
\label{sec:related_work}

\subsection{Spline-based Trajectory Planning in Highway Driving}
\label{subsec:gradient_based_trajectory_planning}
The authors of \cite{gotteSplineBasedMotionPlanning2017} plan the future vehicle motion in a receding horizon manner by solving a nonlinear optimization problem.
The soft constraint problem formulation can be solved efficiently, but may not ensure a collision-free result.
\cite{adajaniaMultiModalModelPredictive2022} solves multiple nonlinear receding horizon problems in parallel to explore different maneuver classes with polynomial splines.
A tailored optimization algorithm solves the problems efficiently.
Though gradient-based approaches can yield high-quality solutions, they only find local minima.
The result from a search over a discretized state space can be the global optimal one but may require many samples.
In \cite{schlechtriemenWigglingComplexTraffic2016} up to 3000 candidate trajectories are generated from which the optimal one is selected.
The prediction of future traffic is used as a heuristic to generate trajectories that are likely to be feasible.
Similarly, \cite{schmidtInteractionAwareLaneChange2019} utilizes traffic prediction to generate many feasible trajectories by applying polygon clipping.
The geometry of the resulting polygons is analyzed to yield trajectories of low cost.

\subsection{Trajectory Planning with Basis Splines}
\label{subsec:trajectory_planning_with_basis_splines}
All the approaches mentioned in section \ref{subsec:gradient_based_trajectory_planning} test the planned trajectories' feasibility at discrete time steps, which fails to ensure time-continuous feasibility.
To deal with this problem, \cite{lengagneGenerationDynamicMotions2010} applies an interval analysis to the planned trajectory.
The extrema of each interval are approximated via the B-spline convex hull.
\cite{vanloockBsplineParameterizedOptimal2015} parameterizes the trajectories of flat systems with B-splines.
The authors exploit that the polynomial, derivative, and integral of a B-spline remains a B-spline.
Thus, all constraints are formulated in B-spline form, and the feasibility can be tested at their respective convex hulls.
In addition, a shrinking horizon problem formulation is proposed and solved via a gradient-based optimization algorithm.
\cite{mercyRealtimeMotionPlanning2016} applies the proposed problem formulation to plan a trajectory for a holonomic mobile robot under consideration of different obstacle shapes.
An extension to non-holonomic robots is proposed in \cite{mercySplineBasedMotionPlanning2017}, which yields a more complex optimization problem than in the holonomic case.
The previous work \cite{dorpmullerOptimizationBSplineParameterized2021} applies the parameterization introduced in \cite{mercySplineBasedMotionPlanning2017} to on-road environments.
In contrast to the current work, a receding horizon problem is formulated that does not guarantee the convergence of the vehicle state into a terminal manifold.\improvement{reference previous publication 1499.1}\improvement{contribution over previous work 8469.1} 
\section{Basis Spline Background}
\label{sec:basis_splines}
This section's content is based on \cite{boorPracticalGuideSplines2001}, if not stated otherwise.
A B-spline $\Spline:\Map{[0,1]}{\Real}$ of order $\OrderBasis \in \NatPos$ is defined by the linear combination of its coefficients (CF) $\Coefs=\RowVect{\Coef_w}\Tp\in\Real^{\DimBasisSet}$ and basis functions $\Basis_w:\Map{[0,1]}{[0,1]}$ with $w \in \CtrlPtsIdxSet = \{0, 1, \ldots, \DimBasisSet - 1\}$ and $\DimBasisSet \in \NatPos$.
\begin{align}
    \Spline(\Inp) = \sum_{l=0}^{\DimBasisSet - 1} \Coef_l \Basis_l(\Inp)
    \label{eq:bspline}
\end{align}
A recursive formula defines the splines' basis from piecewise polynomials over the knot vector $\Brks$.
\begin{align}
    \Brks = \RowVect{\Knot_0 \UnitVect{\OrderBasis}\Tp & \Knot_1 \UnitVect{\OrderBasis - \Cont_1}\Tp & \ldots & \Knot_\Nknots \UnitVect{\OrderBasis - \Cont_\Nknots}\Tp & \Knot_{\Nknots + 1} \UnitVect{\OrderBasis}\Tp}\Tp
    \label{eq:breakpoint}
\end{align}
It is constructed from the BPs $\Knots = \RowVect{\Knot_0 & \Knot_l & \Knot_{\Nknots + 1}}\Tp$ with elements $\Knot_l \in [0,1]$ and $l \in \KnotIdxSet$.
$\Nknots \in \NatPos$ denotes the number of interior BPs and $\KnotIdxSet = \{1, 2, \ldots, \Nknots\}$ are the corresponding indices.
$\Conts = \RowVect{\Cont_l}\Tp$ with elements $\Cont_l \in [0, 1, \ldots, p - 1]$ specifies the order of continuity at the BPs.
The vector $\KnotIvals = \RowVect{\KnotIval_\gamma}\Tp$ contains the BP intervals $\KnotIval_\gamma = \Knot_{\gamma + 1} - \Knot_\gamma$ with $\gamma = 0, 1, \ldots, \Nknots$.
The BP repetition $\UnitVect{u} = \RowVect{1 & 1 & \ldots & 1}\Tp$ with $u = \dim(\UnitVect{u})$ depends on the spline order and the continuity at the corresponding BP.
From now on, it is assumed for all spline bases that $\Knot_0 = 0$.

Each $\Coef_n$ is associated with a Greville site $\Gab_n \coloneqq \frac{1}{\OrderBasis - 1} \sum_{\iota=1}^{\OrderBasis - 1} \Brk_{n + \iota}$.
The control points $\CtrlPts_n = (\Coef_n, \Gab_n)$ with $n \in \CtrlPtsIdxSet$ form a control polygon that encloses the spline function in a convex hull.

The integral and the derivative of a B-spline remains a B-spline function.
In the case of integration, the order is increased, while it is decreased in the case of derivation.
Afterward, a transformation of the CFs into the new basis is performed.
Similarly, the sum or product of two splines $\Spline_\mathrm{a}$ and $\Spline_\mathrm{b}$ remains a B-spline $\Spline_\mathrm{ab}$ \cite{vanloockBsplineParameterizedOptimal2015}.
A combination of the knot vectors $\Brks_\mathrm{a}$ and $\Brks_\mathrm{b}$ is required to form the new basis.
\cite{vanloockBsplineParameterizedOptimal2015} details the rules for combination.
The BP combination is simplified by the assumption that $\Spline_\mathrm{a}$ and $\Spline_\mathrm{b}$ share the same continuities.
In addition, the interior BPs are unique ${\Knot_\mathrm{a}}_n \neq {\Knot_\mathrm{b}}_l, \forall n \in \KnotIdxSet_\mathrm{a} \text{ and } \forall l \in \KnotIdxSet_\mathrm{b}$.
Thus, the sorted, strictly increasing union $\Knots_\mathrm{ab} = \Knots_\mathrm{a} \underset{<}{\cup} \Knots_\mathrm{b}$ of the BP vectors provides the result spline's BPs $\Knots_\mathrm{ab}$.
The new continuities $\Conts_\mathrm{ab} = \Conts_\mathrm{a} \cup \Conts_\mathrm{b}$ are the union of the previous ones.
After the new basis is constructed, the CF transformation matrices can be found by solving a linear equation system.
\section{Problem Formulation}
\label{sec:problem_formulation}
The goal is to formulate a nonlinear optimization problem that yields a solution trajectory ending in the selected terminal manifold.
The terminal manifold comprises a longitudinal and a lateral component that may be reached at different times.
In addition, the recursive feasibility of solutions shall be ensured.
Since the requirements cannot be fulfilled with a single BP scaling, the BP intervals are optimized.
As a result and in contrast to the problem formulation in \cite{vanloockBsplineParameterizedOptimal2015}, the basis functions and their transformations must be differentiated with respect to the BPs.
The spline transformations from spline sums and products are embedded into the nonlinear optimization problem via equality constraints to solve the problem efficiently.
To determine all CFs, the spline functions in (\ref{eq:spline_sum_prod}) must coincide at the Greville sites' indices $u \in \CtrlPtsIdxSet^\times$ and $n \in \CtrlPtsIdxSet^+$ of the sum spline $\Spline^+$ and product spline $\Spline^\times$.
\begin{equation}
    \label{eq:spline_sum_prod}
    \begin{aligned}
        \Spline^+(\Gab^+_n)           & = {\Spline_a}(\Gab^+_n) + {\Spline_b}(\Gab^+_n)               \\
        \Spline^\times(\Gab^\times_u) & = {\Spline_a}(\Gab^\times_u) \cdot {\Spline_b}(\Gab^\times_u)
    \end{aligned}
\end{equation}

\subsection{Trajectory Representation}
\label{subsec:trajectory_representation}
The vehicle trajectory $\Path{\StateVect} = \RowVect{\Path{\StateVectX} & \Path{\StateVectY}}\Tp:\Map{[0,\LenTjy]}{\Real^6}$ of length $\LenTjy \in \RealPos$ spans a 6-dimensional state space.
It results from the concatenation of the trajectory in longitudinal $\Path{\StateVectX}:\Map{[0,\LenTjy]}{\Real^3}$ and lateral direction $\Path{\StateVectY}:\Map{[0,\LenTjy]}{\Real^3}$.
The prescript $\prescript{}{\mathrm{R}}{\square}$ indicates the representation of a state in the target lane center curvilinear coordinate system with the variable placeholder $\square$.
The state in either direction $\Path{\StateVect_i}(t) = \RowVect{\Path{\Pos}_i(t) & \Path{\Vel}_i(t) & \Path{\Acc}_i(t)}\Tp$ contains the vehicle position $\Path{\Pos}_i(t)$, the velocity $\Path{\Vel}_i(t)$ and the acceleration $\Path{\Acc}_i(t)$ in the target lane center reference frame.
In the following, the subscript $i \in [\DirLon, \DirLat]$ indicates the associated direction of motion.
As in \cite{kunzHybridDiscreteparametricOptimization2016}, the vehicle motion is described by a linear differential equation.
Two polynomials of degree five represent a minimum squared jerk position trajectory.
This motivates the utilization of two B-splines $\Spline_{\Pos_\DirLon}$ and $\Spline_{\Pos_\DirLat}$ of order six.
The selection of their continuity vectors ${\Cont_w}_{\Pos_i} = 3$ with $w \in \KnotIdxSet_{\Pos_i}$ ensures the BPs continuity up to the acceleration.\improvement{removed not required text}
Additional B-splines describe the velocity ${\Spline}_{\Vel_i}$, acceleration ${\Spline}_{\Acc_i}$ and jerk ${\Spline}_{\Jerk_i}$ in each direction.
The spline representation of the vehicle state at the normalized time $\TimeNorm \in [0, 1]$ yields $\SplineState_i(\TimeNorm) = \RowVect{\SplinePos{i}(\TimeNorm) & \SplineVel{i}(\TimeNorm) & \SplineAcc{i}(\TimeNorm)}\Tp$.
The CFs of the position $\Coefs_{\Pos_i}$, velocity $\Coefs_{\Vel_i}$, acceleration $\Coefs_{\Acc_i}$ and jerk splines $\Coefs_{\Jerk_i}$ are coupled with the expressions
\begin{equation}
    \ConstrBasis_{i} \coloneqq \RowVect{\TfCoef{\Pos_i}{\Vel_i}     \Coefs_{\Pos_i} - \Coefs_{\Vel_i} \\ \TfCoef{\Vel_i}{\Acc_i}    \Coefs_{\Vel_i} - \Coefs_{\Acc_i} \\ \TfCoef{\Acc_i}{\Jerk_i} \Coefs_{\Acc_i} - \Coefs_{\Jerk_i}}
\end{equation}
and are concatenated in the vector $\CoefsOpt_{i} = \RowVect{\Coefs_{\Pos_i}\Tp & \Coefs_{\Vel_i}\Tp & \Coefs_{\Acc_i}\Tp & \Coefs_{\Jerk_i}\Tp}\Tp$.
$\TfCoef{\mathrm{a}}{\mathrm{b}}$ denotes the CF transformation matrix from spline $\Spline_\mathrm{a}$ to $\Spline_\mathrm{b}$ as a result of the time derivative.
The splines' BPs are defined on a common vector $\Knots_{\DirLon\DirLat}$.
The interior BPs of the longitudinal splines $\Knots_\DirLon = \RowVect{{\Knot_{\DirLon\DirLat}}_0 & {\Knot_{\DirLon\DirLat}}_l & {\Knot_{\DirLon\DirLat}}_{\Nknots_{\DirLon\DirLat} + 1}}\Tp$ are composed of a subset of the common interior BPs $l \in \KnotIdxSet_\DirLon \subseteq \KnotIdxSet_{\DirLon\DirLat}$.
The remaining common BPs are assigned to the BPs of the lateral splines $\Knots_\DirLat = \RowVect{{\Knot_{\DirLon\DirLat}}_0 & {\Knot_{\DirLon\DirLat}}_n & {\Knot_{\DirLon\DirLat}}_{\Nknots_{\DirLon\DirLat} + 1}}\Tp$ with $n \in \KnotIdxSet_\DirLat = \KnotIdxSet_{\DirLon\DirLat} \setminus \KnotIdxSet_\DirLon$.
Both BP vectors share the same start and end BPs.

\subsection{Other Vehicle Consideration}
\label{subsec:other_vehicle_consideration}
It is assumed that rectangles accurately describe vehicle shapes.
Similar to \cite{adajaniaMultiModalModelPredictive2022}, the rectangles are overapproximated by axis-aligned ellipses.
The ellipses are centered in the vehicles' center of mass with diameters $\AxisLen{i} \in \RealPos$ in x- and y-direction, respectively for the ego vehicle.
Similarly, the diameters $\AxisLenOv{i} \in \RealPos$ are defined for all $\NumOv \in \NatPos$ obstacle vehicles with $m = 1, 2, \ldots, \NumOv$.\improvement{removed ellipse equations}
$\PosPathOv{i}(\Time) \in \Real$ denotes the $i$-position of the $m$-th obstacle vehicle.
The ellipse encloses the rectangular vehicle shape under the assumption of a low heading angle deviation from the reference path.
Since the ellipses are axis aligned, the Minkowski sum of the ego vehicle and the obstacles shapes reduce to a sum over the ellipse parameters.
Thus, an overlap of vehicle shapes does not occur if the constraint (\ref{eq:constr_veh_shape}) holds.
\begin{equation}
    \label{eq:constr_veh_shape}
    \frac{\big(\PosPathX(\Time) - \PosPathXov(\Time)\big)^2}{(\AxisLenX + \AxisLenXov)^2} + \frac{\big(\PosPathY(\Time) - \PosPathYov(\Time)\big)^2}{(\AxisLenY + \AxisLenYov)^2}  \geq \frac{1}{4}
\end{equation}\improvement{corrected constraint expression}
To enable a B-spline formulation of the constraints, the other vehicle trajectories are described by B-splines $\SplinePosXov$ and $\SplinePosYov$ of order six without any interior BPs.
To formulate constraint (\ref{eq:constr_veh_shape}) with B-spline CFs, several intermediate B-splines are introduced to perform the required arithmetic operations.
Since the obstacle trajectories have a fixed length of $\SI{10}{\second}$, they must be first transformed to variable length representations $\SplinePosOvVar{\DirLon}$ and $\SplinePosOvVar{\DirLat}$.
\begin{equation}
    \label{eq:constr_basis_pos_ov}
    \ConstrBasisPosOvVar{i} \coloneqq \RowVect{{\SplineGabPosOvVar{i}}_w - \SplinePosOv{i}\left({\GabsPosOvVar{i}}_w\right)}\Tp
\end{equation}
${\SplineGabPh}_\gamma \coloneqq \SplinePh({\GabPh}_\gamma)$ evaluates a spline at its Greville site ${\GabPh}_\gamma$ with $\gamma \in \CtrlPtsIdxSet_{\square}$.
The equality constraints $\ConstrBasisPosOvVar{i}$ ensure that the CFs of the variable length spline $\Coefs_{\tilde{\Pos}_{i}}$ are chosen to ensure the coincidence between $\SplinePosOvVar{i}$ and $\SplinePosOv{i}$ at the Greviell sites ${\GabsPosOvVar{i}}_w$ with the indices $w \in \CtrlPtsIdxSet_{\PosOvVar{i}}$.
Note that the dimensionality of $\ConstrBasis_\square \in \Real^{\dim(\CtrlPtsIdxSet_\square)}$ grows with the number of Greville sites.
Next, the spline representation for the position difference $\SplinePosDiffOv{i} = \SplinePos{i} - \SplinePosOvVar{i}$ between ego vehicle and obstacle vehicle trajectories are calculated.
Equation (\ref{eq:constr_basis_pos_diff_ov}) couples the CFs of $\SplinePos{i}$ and $\SplinePosOvVar{i}$ to the CFs of $\SplinePosDiffOv{i}$ with $u \in \CtrlPtsIdxSet_{\PosDiffOv{i}}$.
Then, the squared position difference $\SplinePosDiffSqOv{i} = \SplinePosDiffOv{i} \cdot \SplinePosDiffOv{i}$ is calculated.
Equation (\ref{eq:constr_basis_pos_diff_sq_ov}) couples the respective spline coeffcients with $n \in \CtrlPtsIdxSet_{\PosDiffSqOv{i}}$.
\begin{align}
    \label{eq:constr_basis_pos_diff_ov}
     & \ConstrBasisPosDiffOv{i} \coloneqq \left[\SplinePos{i}\left({\GabPosDiffOv{i}}_u\right) - \SplinePosOvVar{i}\left({\GabPosDiffOv{i}}_u\right) - {\SplineGabPosDiffOv{i}}_u\right]\Tp \\
    \label{eq:constr_basis_pos_diff_sq_ov}
     & \ConstrBasisPosDiffSqOv{i} \coloneqq \RowVect{\bigg(\SplinePosDiffOv{i}\left({\GabPosDiffSqOv{i}}_n\right)\bigg)^2 - {\SplineGabPosDiffSqOv{i}}_n}\Tp
\end{align}\improvement{more compact equation}
Finally, $\SplineEllipseOv$ implements the collision avoidance constraint via the expression
\begin{equation}
    \label{eq:constr_veh_shape_spline}
    \ConstrBasisEllipseOv \coloneqq \left[\frac{\SplinePosDiffSqOv{\DirLon}\left({\GabEllipseOv}_\iota\right)}{(\AxisLenX + \AxisLenXov)^2} + \frac{\SplinePosDiffSqOv{\DirLat}\left({\GabEllipseOv}_\iota\right)}{(\AxisLenY + \AxisLenYov)^2} - {\SplineGabEllipseOvQ}_\iota\right]\Tp
\end{equation}\improvement{corrected constraint expression}
with the index $\iota \in \CtrlPtsIdxSet_{\EllipseOv}$ and $\SplineEllipseOvQ = \SplineEllipseOv + \frac{1}{4}$.
The vectors $\CoefsOpt_{\mathrm{o}_m} = \RowVect{\Coefs_{\tilde{\Pos}_{i}}\Tp & \Coefs_{\PosDiffOv{i}}\Tp & \Coefs_{\PosDiffSqOv{i}}\Tp & \Coefs_{\EllipseOv}\Tp}\Tp$ and $\ConstrBasis_{\mathrm{o}_m} = \RowVect{\ConstrBasisPosOvVar{i}\Tp & \ConstrBasisPosDiffOv{i}\Tp & \ConstrBasisPosDiffSqOv{i}\Tp & \ConstrBasisEllipseOv\Tp}\Tp$ contain the respective CFs and constraints related to the other vehicles.

\subsection{Heading Angle Limitation}
\label{subsec:heading_angle_limitation}
To ensure that the low heading angle assumption is not violated, the heading angle of the ego vehicle is limited.
According to \cite{rathgeberTrajektorienplanungUndFolgeregelung2016}, the path relative heading angle $\HeadPath(\Time) \in \Real$ depends on the vehicle state and the path curvature $\CurvPath(\Time) \in \Real$.
\begin{align}
    \tan \HeadPath(\Time) = \frac{\VelPathY(\Time)}{\VelPathX(\Time) \big(1 - \CurvPath(\Time) \PosPathY(\Time)\big)}
    \label{eq:heading}
\end{align}
A B-spline formulation of the constraints (\ref{equ:heading_bound}) with the relative heading bound $\HeadBound \in \RealPos$ is enabled by introducing the constraint expressions (\ref{equ:heading_ubnd}) and (\ref{eq:heading_lbnd}) for the upper and the lower bound, respectively.
Rational expressions are retained by introducing $\HeadBound_\mathrm{t} = \tan \HeadBound$.
\begin{align}
    \label{equ:heading_bound}
    - \HeadBound_\mathrm{t} & \leq \tan \HeadPath(\Time) \leq \HeadBound_\mathrm{t}                                                          \\
    \label{equ:heading_ubnd}
    \Leftrightarrow 0       & \leq \VelPathX(\Time) \big(1 - \CurvPath(\Time) \PosPathY(\Time)\big) \HeadBound_\mathrm{t} - \VelPathY(\Time) \\
    \label{eq:heading_lbnd}
    \Leftrightarrow 0       & \leq \VelPathX(\Time) \big(1 - \CurvPath(\Time) \PosPathY(\Time)\big) \HeadBound_\mathrm{t} + \VelPathY(\Time)
\end{align}
The resulting inequalities are simplified with the maximum assumed road curvature $\CurvPathMax \in \RealPos$ and maximum y-position $\PosYMax \in \RealPos$ in a conservative manner.
The upper and lower bound constraints are reformulated as spline functions $\SplineHeadUb = \SplineVelX \HeadBndFct - \SplineVelY$ and $\SplineHeadLb = \SplineVelX \HeadBndFct + \SplineVelY$ with the factor $\HeadBndFct = (1 - \CurvPathMax \PosYMax) \HeadBound_\mathrm{t}$.
\begin{equation}
    \label{equ:heading_ubnd_spline}
\begin{aligned}
    \ConstrBasisHeadUb &\coloneqq \left[\SplineVelX\left({\GabHeadUb}_n\right) \HeadBndFct - \SplineVelY\left({\GabHeadUb}_n\right) - {\SplineGabHeadLb}_n\right]\Tp\\
    \ConstrBasisHeadLb &\coloneqq \left[\SplineVelX\left({\GabHeadLb}_u\right) \HeadBndFct  + \SplineVelY\left({\GabHeadLb}_u\right) - {\SplineGabHeadLb}_u\right]\Tp
\end{aligned}
\end{equation}
The expressions in (\ref{equ:heading_ubnd_spline}) introduce a relation between the respective spline CFs with $n \in \CtrlPtsIdxSet_{\HeadBound_{\mathrm{ub}}}$ and $u \in \CtrlPtsIdxSet_{\HeadBound_{\mathrm{lb}}}$.

\subsection{Terminal Manifolds}
\label{subsec:terminal_manifolds}
The trajectory shall end in a region of the state space that is assumed to be selected by a behavior planner.
In the lateral direction, the trajectory merges into the center line of the target lane.
A target speed $\VelTarget \in \RealPos$ shall be reached longitudinally.
It must be decided in advance which part of the terminal manifold to reach first.
For each decision, two terminal index sets $\KnotIdxSetTerm{\DirLon}{\lambda}$ and $\KnotIdxSetTerm{\DirLat}{\lambda}$ with $\lambda \in \{\mathrm{lon}, \mathrm{lat}\}$ are introduced to formulate the constraint expressions.
In case $\VelTarget$ is reached first, the index sets $n \in \KnotIdxSetTerm{\DirLon}{\mathrm{lon}} = \{\Nknots_\DirLon, \Nknots_\DirLon + 1\}$ and $u \in \KnotIdxSetTerm{\DirLat}{\mathrm{lon}} = \{\Nknots_\DirLat + 1\}$ define the terminal constraints.
In this configuration, the terminal velocity is reached already at the second last BP ${\Knot_\DirLon}_{\Nknots_\DirLon}$.
If the lane center shall be reached first, the index sets $n \in \KnotIdxSetTerm{\DirLon}{\mathrm{lat}} = \{\Nknots_\DirLon + 1\}$ and $u \in \KnotIdxSetTerm{\DirLat}{\mathrm{lat}} = \{\Nknots_\DirLat, \Nknots_\DirLat + 1\}$ are used.
The equations (\ref{eq:terminal_constraints}) formulate the terminal constraints.
\begin{equation}
    \label{eq:terminal_constraints}
    \begin{aligned}
        \ConstrTerm{\DirLon}{\lambda} & \coloneqq \RowVect{\SplineVel{\DirLon}({\Knot_\DirLon}_n) - \VelTarget & \quad \SplineAcc{\DirLon}({\Knot_\DirLon}_n)}\Tp                                              \\
        \ConstrTerm{\DirLat}{\lambda} & \coloneqq \RowVect{\SplinePos{\DirLat}({\Knot_\DirLat}_u)              & \quad \SplineVel{\DirLat}({\Knot_\DirLat}_u)     & \quad \SplineAcc{\DirLat}({\Knot_\DirLat}_u)}\Tp
    \end{aligned}
\end{equation}

\subsection{Cost Function}
\label{subsec:cost_function}
The final trajectory shall compromise comfort and progress toward the selected terminal manifold.
As in \cite{kunzHybridDiscreteparametricOptimization2016}, the cost is defined as a weighted sum of the time and the integral of the squared jerk $\prescript{}{\mathrm{R}}{j^2_i}(t)$ into the longitudinal and lateral terminal manifold.
Therefore, the spline function $\SplineJerkSq{i} = \SplineJerk{i} \cdot \SplineJerk{i}$ represents the squared jerk in the $i$-direction.
According to section (\ref{subsec:trajectory_representation}), the BP continuity is chosen up to the second derivative.
Thus, the jerk trajectory is $C^{-1}$ continuous.
The CFs at the discontinuous BPs and the initial and final BP coincide with the spline function and can be compared directly with $w \in \CtrlPtsIdxSet_{\JerkSq{i}}$.
\begin{equation}
    \label{eq:jerk_squared}
    \ConstrBasisJerkSq{i} \coloneqq \RowVect{{\Coef_{\JerkSq{i}}}_w - {\Coef^2_{\Jerk_{i}}}_w}\Tp
\end{equation}

\subsection{Nonlinear Optimization Problem}
\label{subsec:nonlinear_optimization_problem}
Finally, the introduced expressions are used to formulate the nonlinear optimization problem (\ref{eq:optimization_problem}).
The CFs of all splines subject to optimization are accumulated in the vector $\CoefsOpt = \RowVect{\CoefsOpt_{i}\Tp & \Coefs_{\JerkSq{i}}\Tp & \CoefsOpt_{\mathrm{o}_m}\Tp & \Coefs_{\HeadBound_{\mathrm{lb}}}\Tp  & \Coefs_{\HeadBound_{\mathrm{ub}}}\Tp}\Tp$
and the constraint expressions in $\ConstrBasis = \RowVect{\ConstrBasis_{i}\Tp & \ConstrBasisJerkSq{i}\Tp & \ConstrBasis_{\mathrm{o}_m}\Tp & \ConstrBasisHeadLb\Tp & \ConstrBasisHeadUb\Tp}\Tp$.
To ensure the linear independence of the constraint expressions, the BP intervals $\KnotIvals_{\DirLon\DirLat}$ must be lower bounded by $\KnotIvalMin \in \RealPos$.
The constraint $\sum_{\iota=0}^{\Nknots_{\DirLon \DirLat}} {\KnotIval_{\DirLon\DirLat}}_\iota \leq 1$ copes for the limited prediction horizon.
Dynamic feasibility is ensured with constraints on the CFs of $\SplineAccX$ and $\SplineAccY$.
Therefore, the approximations from \cite{rathgeberTrajektorienplanungUndFolgeregelung2016} for the vehicle's longitudinal and lateral accelerations $\Norm{\Acc}_\DirLon(\Time) \approx \Path{\Acc}_\DirLon(\Time)$ and $\Norm{\Acc}_\DirLat(\Time) \approx \Path{\Acc}_\DirLat(\Time) - \CurvPath(\Time) \Path{\Vel}^2_\DirLon(\Time)$.
The constraints are simplified with the maximum velocity $\VelBound \in \RealPos$ and maximum road curvature $\CurvPathMax$.
The prescript $\prescript{}{\mathrm{N}}{\square}$ indicates the definition of a state value in the ego vehicle's normalized coordinate system.
The weights $\omega_{T_i} \in \RealPos$ and $\omega_{j_i} \in \RealPos$ balance progress and comfort in each direction.
\begin{align}
    \label{eq:optimization_problem}
    \begin{aligned}
         & \min_{\CoefsOpt, \KnotIvals_{\DirLon\DirLat}} \sum_{n \in \{\DirLon, \DirLat\}} \sum_{u \in \KnotIdxSetTerm{n}{\lambda}} \omega_{T_n}{\Knot_n}_u+ \int_{0}^{1} \omega_{j_n} \SplineJerkSq{n}(\tau)\mathrm{d}\tau \\
         & \text{such that}                                                                                                                                                                                                 \\
         & \begin{matrix}\SplineState_i(0) = \StateVect_i(0), & \ConstrBasis = 0, & \ConstrTerm{\DirLon}{\lambda} = 0, & \ConstrTerm{\DirLat}{\lambda} = 0,\end{matrix}                                                                                                                                                                                       \\
         & \begin{matrix}\Coefs_{\EllipseOv} \geq 0, & \Coefs_{\HeadBound_{\mathrm{lb}}} \geq 0, & \Coefs_{\HeadBound_{\mathrm{ub}}} \geq 0, & \Coefs_{\VelX} \leq \VelBound,\end{matrix}                                                                                                                                                                                       \\
         & \begin{matrix}\KnotIvals_{\DirLon\DirLat} \geq \KnotIvalMin, & \sum_{\iota=0}^{\Nknots_{\DirLon \DirLat}} {\KnotIval_{\DirLon\DirLat}}_\iota \leq 1,\end{matrix}                                                                                                                                                                                       \\
         & \begin{matrix}\AccBound_\DirLon \geq \Coefs_{\Acc_\DirLon} \geq -\AccBound_\DirLon, & \AccBound_\DirLat - \CurvPathMax \VelBound^2 \geq \Coefs_{\Acc_\DirLat} \geq -\AccBound_\DirLat + \CurvPathMax \VelBound^2\end{matrix}                                                                                                                                                                                       \\
    \end{aligned}
\end{align}
The optimization problem is implemented with CasADi \cite{anderssonCasADiSoftwareFramework2019} and solved with IPOPT \cite{wachterImplementationInteriorpointFilter2006}. 
\section{Breakpoint Adaption}
\label{sec:breakpoint_adaption}
\graphicspath{{figures/breakpoint_adaption/}}

\begin{figure}[tb]
    \vspace{5pt}
    \centering
    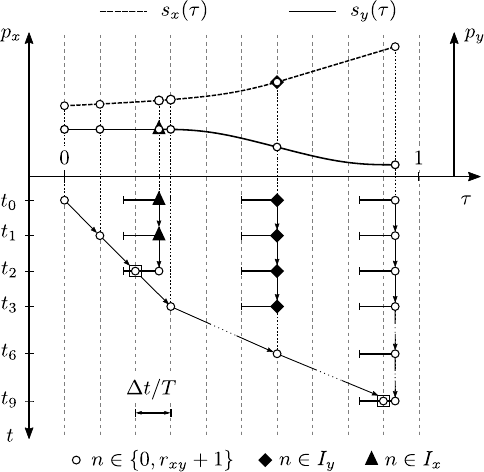
    \setlength{\belowcaptionskip}{-10pt}
    \caption{Exemplary visualization of the BP evolution in closed-loop.
    Each marker in the diagram represents a BP ${\Knot_{\DirLon\DirLat}}_n$ from the common BP vector.
    Markers enclosed by a square are not an element in $\Knots_{\DirLon\DirLat}$.}
    \label{fig:breakpoint_adaption}
\end{figure}

Since the planning algorithm is implemented in a shrinking horizon manner, the intervals between the BPs get smaller during closed-loop.
The problem formulation (\ref{eq:optimization_problem}) demands a minimum distance between two BPs and therefore requires the removal of BPs during closed-loop simulation.
Fig. \ref{fig:breakpoint_adaption} illustrates the evolution of an exemplary common BP vector.
In the following, it is assumed that the future motion of all vehicles is known and the ground truth states of all objects are available.

In this example, there exists a total number of 4 BPs at time step $\Time_0$.
The first intermediate BP is assigned to the x-direction, and the second to the y-direction.
Since all BPs are subject to optimization, they can be arranged to reproduce the remaining part of the previous solution that is still feasible in the next step.
Though this solution is not always optimal, this property ensures recursive feasibility.
$\Delta t \in \RealPos$ denotes the simulation time increment.
According to \cite{boorPracticalGuideSplines2001}, a refinement of the knot vector decreases the conservatism of the B-spline convex hull.
Thus, the optimization algorithm might find a better solution in active inequality constraint cases.

In time step $\Time_2$, the first BP is moved closer to the second one, violating the minimum distance constraint.
The problem does not consider the first BP (\ref{eq:optimization_problem}) to retain a feasible solution.
Instead, the former second BP is now the first one of the common BP vector.

The horizon shortens further in time step $\Time_3$, and the first BP is removed.
This strategy can be applied until time step $\Time_{9}$.
No degrees of freedom are left, and the vehicle follows the remaining trajectory into the terminal manifold.\todo{reinforce difference to related work!}
\section{Evaluation}
\label{sec:evaluation}
\graphicspath{figures/evaluation/}

\begin{figure}[tb]
  \centering
  \vspace{5pt}
  \input{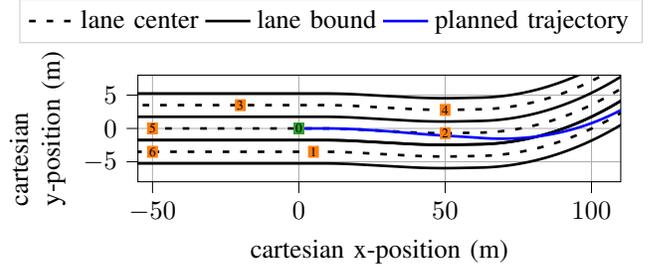}
  \setlength{\abovecaptionskip}{-10pt}
  \setlength{\belowcaptionskip}{-10pt}
  \caption{Scene at $t = \SI{0}{\second}$ from the ego vehicle perspective.
  The ego vehicle is colored green, and the other ones are orange.
  The planned trajectory consists of 3 lateral BPs.}
  \label{fig:scene_0_ego}
\end{figure}

The evaluation is performed on a highway section of Town04 in Carla \cite{dosovitskiyCARLAOpenUrban2017} with up to five obstacle vehicles.
Fig. \ref{fig:scene_0_ego} shows the initial scene from the ego vehicle perspective.
All vehicles are marked with an identifier, where 0 denotes the ego vehicle.
The ego vehicle starts on a straight road segment that connects to a curved one with $\CurvPathMax = \SI[parse-numbers=false]{1/180}{\per\meter}$.
Initially, it drives with a velocity of $\SI{50}{\kilo\meter\per\hour}$ and shall accelerate to $\VelTarget=\SI{70}{\kilo\meter\per\hour}$.
At the same time, it performs a lane change to the right.
All obstacle vehicles drive with a constant velocity along their initial lane centers.
Vehicles 1, 4 and 5 keep a velocity of $\SI{50}{\kilo\meter\per\hour}$, while vehicles 2 and 3 drive with $\SI{70}{\kilo\meter\per\hour}$.

Similar to \cite{adajaniaMultiModalModelPredictive2022}, the vehicles are overapproximated with axis-aligned ellipses assuming a maximum heading deviation of $\SI{13}{\deg}$ for other vehicles and $\HeadBound = \SI{10}{\deg}$ for the ego vehicle.
The planned trajectory is scaled to $\LenTjy = \SI{10}{\second}$.
An exact realization is assumed at a simulation time increment of $\Delta t=\SI{0.1}{\second}$ so the theoretical properties can be demonstrated that only hold in the nominal case.\improvement{clarify application issues in the presence of disturbances 1499.5, 1501.2}
Thus, the closed-loop trajectory is formed by setting the vehicle to the next planned state along the open-loop trajectory.\improvement{clarify open-loop and closed-loop trajectory 8469.4}
In the following, the target velocity shall be reached first ($\lambda=\mathrm{lon}$), and the number of inner BPs in the x-direction is fixed to $\Nknots_x = 1$.
The initial guess is based on two polynomials of degree five, which connect the current state with the target velocity in \SI{7}{\second} and the target lane in \SI{9}{\second}.
A predetermined number of BPs is distributed equidistantly along the polynomials to form the equivalent B-splines.\improvement{detail breakpoint selection 1499.4}
The initial trajectory would lead to a collision with the vehicle in the right lane and is infeasible.

\begin{figure}[b]
  \centering
  \vspace{-10pt}
  \begin{subfigure}[t]{0.45\textwidth}
    \input{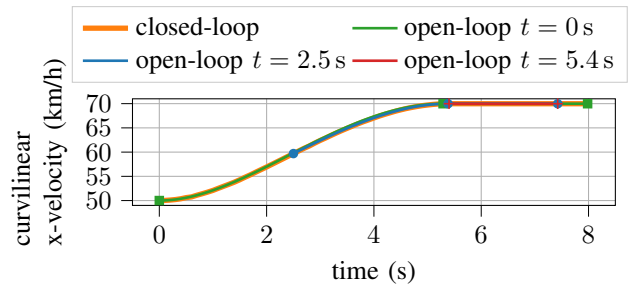}
    \setlength{\abovecaptionskip}{-10pt}
    \caption{Closed- and open-loop velocity trajectories in the x-direction.}
    \label{fig:vx_kx3_ky3}
  \end{subfigure}
  \begin{subfigure}[t]{0.45\textwidth}
    \input{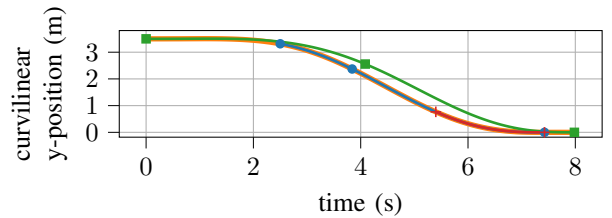}
    \setlength{\abovecaptionskip}{-10pt}
    \caption{Closed- and open-loop y-position trajectories.}
    \label{fig:y_kx3_ky3}
  \end{subfigure}
  \caption{Open-loop and closed-loop trajectories result from the planner configuration with $\Nknots_\DirLon=1$ and $\Nknots_\DirLat=1$.
  The spline BPs are indicated by the markers.}
  \vspace{-3.2pt}
  \label{fig:kx3_ky3}
\end{figure}

Fig. \ref{fig:kx3_ky3} shows the closed-loop velocity in the x-direction and the y-position in the target lane frame together with three planned trajectories.
The vehicle converges asymptotically into the terminal manifold.
One can observe that the closed-loop trajectory deviates from the initially planned one.
The deviation is the result of the decreasing conservativeness of the B-spline hull.
During the simulation, the distance of the convex hulls from the splines decrease.
Thus, active constraints become less conservative during simulation in the presence of the obstacle in the right lane.
Starting from $t = \SI{2.5}{\second}$, the obstacle vehicle 1 has been passed, and no inequality constraint is active anymore.
As a result, the planned trajectory coincides with the closed-loop trajectory.
The ego vehicle reaches its desired velocity at $t = \SI{5.4}{\second}$.
The intermediate BPs in the common BP vector are removed at this time.

\begin{figure}[tb]
  \centering
  \begin{tikzpicture}

\definecolor{crimson2143940}{RGB}{214,39,40}
\definecolor{darkgray176}{RGB}{176,176,176}
\definecolor{darkorange25512714}{RGB}{255,127,14}
\definecolor{forestgreen4416044}{RGB}{44,160,44}
\definecolor{lightgray204}{RGB}{204,204,204}
\definecolor{mediumpurple148103189}{RGB}{148,103,189}
\definecolor{sienna1408675}{RGB}{140,86,75}
\definecolor{steelblue31119180}{RGB}{31,119,180}

\begin{axis}[
legend cell align={left},
legend style={at={(1,1)}, fill opacity=0.8, draw opacity=1, text opacity=1, draw=lightgray204},
tick align=outside,
tick pos=left,
x grid style={darkgray176},
xmajorgrids,
xmin=-0.5, xmax=8.5,
xtick style={color=black},
xlabel={simulation time (s)},
y grid style={darkgray176},
ymajorgrids,
ymin=-0.130161859747022, ymax=2.2,
ytick style={color=black},
ylabel={objective value},
width=0.45\textwidth,
height=3.3cm,
]
\addplot [semithick, steelblue31119180, mark=*, mark size=0.5, mark repeat=4, mark options={solid}, line width=1pt]
table {%
0.0 2.05045771598816
0.1 1.96690785884857
0.2 1.88980305194855
0.3 1.8151615858078
0.4 1.73791992664337
0.5 1.66039884090424
0.6 1.60129356384277
0.7 1.54697704315186
0.8 1.4971045255661
0.9 1.45039582252502
1.0 1.4013683795929
1.1 1.34717726707458
1.2 1.29029631614685
1.3 1.24165725708008
1.4 1.20162546634674
1.5 1.16744661331177
1.6 1.13635814189911
1.7 1.10634171962738
1.8 1.0733460187912
1.9 1.03805637359619
2.0 1.00325000286102
2.1 0.973940134048462
2.2 0.949010610580444
2.3 0.924488186836243
2.4 0.900111615657806
2.5 0.875758111476898
2.6 0.851320207118988
2.7 0.826705157756805
2.8 0.801833868026733
2.9 0.776640295982361
3.0 0.751070320606232
3.1 0.725081384181976
3.2 0.698641240596771
3.3 0.671727299690247
3.4 0.644325852394104
3.5 0.616431176662445
3.6 0.588044762611389
3.7 0.559174358844757
3.8 0.529833316802979
3.9 0.50003969669342
4.0 0.469815492630005
4.1 0.439185678958893
4.2 0.408177495002747
4.3 0.376819610595703
4.4 0.34514132142067
4.5 0.307286620140076
4.6 0.296997666358948
4.7 0.284977734088898
4.8 0.273097693920135
4.9 0.261383324861526
5.0 0.249856039881706
5.1 0.238532066345215
5.2 0.227421671152115
5.3 0.21652826666832
5.4 0.205847650766373
5.5 0.195367127656937
5.6 0.185064762830734
5.7 0.174908459186554
5.8 0.164855197072029
5.9 0.154850199818611
6.0 0.144826129078865
6.1 0.134702190756798
6.2 0.124383419752121
6.3 0.113759756088257
6.4 0.102705284953117
6.5 0.0910773649811745
6.6 0.0787158608436584
6.7 0.0654422789812088
6.8 0.0510589405894279
6.9 0.0353481881320477
7.0 0.0
7.1 0.0
7.2 0.0
7.3 0.0
7.4 0.0
7.5 0.0
7.6 0.0
7.7 0.0
7.8 0.0
7.9 0.0
8.0 0.0
};
\addlegendentry{$\Nknots_\DirLat = 0$}
\addplot [semithick, darkorange25512714, mark options={solid}, line width=1pt]
table {%
0.0 1.65844297409058
0.1 1.61963224411011
0.2 1.58263826370239
0.3 1.54679882526398
0.4 1.51177752017975
0.5 1.47743022441864
0.6 1.44357490539551
0.7 1.41092216968536
0.8 1.3795919418335
0.9 1.34904289245605
1.0 1.31946980953217
1.1 1.290731549263
1.2 1.26271629333496
1.3 1.23582649230957
1.4 1.21026074886322
1.5 1.18580806255341
1.6 1.16185939311981
1.7 1.13824677467346
1.8 1.11499416828156
1.9 1.09213006496429
2.0 1.06962668895721
2.1 1.04748296737671
2.2 1.02571165561676
2.3 1.0043625831604
2.4 0.983278810977936
2.5 0.96256297826767
2.6 0.942127466201782
2.7 0.921841979026794
2.8 0.901590466499329
2.9 0.881270289421082
3.0 0.860791683197021
3.1 0.840076684951782
3.2 0.819058775901794
3.3 0.797681927680969
3.4 0.775899708271027
3.5 0.753674924373627
3.6 0.730978608131409
3.7 0.707789242267609
3.8 0.684092283248901
3.9 0.659879088401794
4.0 0.63514643907547
4.1 0.609895586967468
4.2 0.584131717681885
4.3 0.557863056659698
4.4 0.531100273132324
4.5 0.503855466842651
4.6 0.476141661405563
4.7 0.447972118854523
4.8 0.419359356164932
4.9 0.39031457901001
5.0 0.360846817493439
5.1 0.330962330102921
5.2 0.286467462778091
5.3 0.245572075247765
5.4 0.242405384778976
5.5 0.231280073523521
5.6 0.220374688506126
5.7 0.2096838504076
5.8 0.199194118380547
5.9 0.188883289694786
6.0 0.178719624876976
6.1 0.168661087751389
6.2 0.158654659986496
6.3 0.148635521531105
6.4 0.138526394963264
6.5 0.128236711025238
6.6 0.117661967873573
6.7 0.106682881712914
6.8 0.0951647311449051
6.9 0.0829565674066544
7.0 0.069890484213829
7.1 0.0557808876037598
7.2 0.040423721075058
7.3 0.0235957335680723
7.4 0.0
7.5 0.0
7.6 0.0
7.7 0.0
7.8 0.0
7.9 0.0
8.0 0.0
};
\addlegendentry{$\Nknots_\DirLat = 1$}
\addplot [semithick, dashed, crimson2143940, mark options={solid}, line width=1pt]
table {%
0.0 1.65758407115936
0.1 1.61876583099365
0.2 1.58177256584167
0.3 1.54593050479889
0.4 1.51089668273926
0.5 1.47652590274811
0.6 1.44262838363647
0.7 1.40993297100067
0.8 1.37857472896576
0.9 1.347984790802
1.0 1.31838369369507
1.1 1.28961622714996
1.2 1.26156878471375
1.3 1.23466432094574
1.4 1.20906293392181
1.5 1.18461191654205
1.6 1.16066312789917
1.7 1.1370393037796
1.8 1.11378228664398
1.9 1.09091222286224
2.0 1.0684175491333
2.1 1.04628443717957
2.2 1.02454423904419
2.3 1.00324249267578
2.4 0.982164800167084
2.5 0.961497902870178
2.6 0.930303394794464
2.7 0.920837581157684
2.8 0.900601923465729
2.9 0.880289554595947
3.0 0.859812319278717
3.1 0.839093744754791
3.2 0.818068504333496
3.3 0.796681642532349
3.4 0.774888038635254
3.5 0.752651393413544
3.6 0.729943513870239
3.7 0.706743955612183
3.8 0.683038592338562
3.9 0.658819615840912
4.0 0.634084224700928
4.1 0.608834207057953
4.2 0.583075046539307
4.3 0.556815207004547
4.4 0.530065476894379
4.5 0.502838134765625
4.6 0.475146144628525
4.7 0.447002649307251
4.8 0.418419986963272
4.9 0.38940903544426
5.0 0.359978526830673
5.1 0.330134242773056
5.2 0.287385880947113
5.3 0.245048463344574
5.4 0.241091698408127
5.5 0.22999219596386
5.6 0.219112426042557
5.7 0.208446070551872
5.8 0.19797870516777
5.9 0.187687039375305
6.0 0.177538126707077
6.1 0.16748870909214
6.2 0.157484367489815
6.3 0.147458910942078
6.4 0.137333482503891
6.5 0.127015963196754
6.6 0.116400122642517
6.7 0.105364933609962
6.8 0.0937737971544266
6.9 0.0814738348126411
7.0 0.0682951211929321
7.1 0.0540499500930309
7.2 0.0385320857167244
7.3 0.00999999977648258
7.4 0.0
7.5 0.0
7.6 0.0
7.7 0.0
7.8 0.0
7.9 0.0
8.0 0.0
};
\addlegendentry{$\Nknots_\DirLat = 2$}
\end{axis}

\end{tikzpicture}
  \setlength{\belowcaptionskip}{-10pt}
  \caption{Evolution of the objective function values from different BP vectors over the simulation time.}
  \label{fig:closed_loop}
\end{figure}
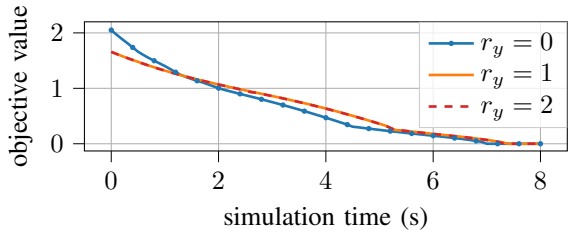

The asymptotic convergence can also be observed in Fig. \ref{fig:closed_loop}, where monotonically decreasing cost function values are displayed for different BP numbers.
An increase in the BP number from $\Nknots_\DirLat=0$ to $\Nknots_\DirLat=1$ decreases the initial cost value.
A single interior BP delays the lane change until the obstacle vehicle is passed, resulting in a lower jerk.
The cost decrease from the two inner BPs is low in comparison.
The two polynomial segments in the lateral direction already result in a low-cost solution such that an additional segment yields only a small improvement.
As soon as one of the two terminal manifolds is reached, the change of the cost values decreases since the cost of one direction constantly remains at zero.
The effect can be observed, e.g., for the configuration with $\Nknots_\DirLat=1$ at $t = \SI{5.4}{\second}$.

The time until IPOPT meets its convergence criteria during the generation of the initial trajectory is evaluated for all configurations over 20 repetitions (Ubuntu 22.04, AMD Ryzen 5 3600 CPU at \SI{3.6}{\giga\hertz}, \SI{16}{\giga\byte} RAM).
At first, only obstacles 1 and 2 are considered.
While it takes a median time of \SI{42}{\milli\second} in case of $\Nknots_\DirLat = 0$ to converge, a lower time of \SI{39}{\milli\second} can be observed for $\Nknots_\DirLat = 1$.
Though the problem complexity is lower with no interior BP, the convex hull is more conservative, too.
In turn, IPOPT starts with a higher initial infeasibility that requires \si{10} more iterations to converge.
Adding another BP increases the median time to \SI{120}{\milli\second}.

In the following, the number of inner breakpoints is fixed to $\Nknots_\DirLat = 1$ while the number of obstacle vehicles increases.
Considering obstacles 3 and 4, in addition to the first two vehicles, increases the median time to \SI{84}{\milli\second}.
In the presence of all six obstacles, it takes a median time of \SI{98}{\milli\second} to solve the optimization problem.
\section{Summary and Outlook}
\label{sec:summary}
This paper proposes a nonlinear optimization problem that ensures the feasibility of the planned trajectory via a B-spline parameterization in highway scenarios.
The planning problem is implemented so the trajectory can end in a target velocity and lane at two different times.
The spline BPs are subject to optimization, and the proposed BP removal strategy enables recursive feasibility.
Implementing basis transformations with additional constraints and optimization variables achieves a sparse optimization problem.

The lane change scenario evaluation shows the vehicle's convergence into the terminal manifold.
Due to the chosen cost function, time-consistent planning results are achieved if no inequality constraint is active.
In addition, the influence of the BP number in the lateral direction on the optimal solution is evaluated.

Next, a leading vehicle's predicted trajectory shall be considered a terminal manifold in the optimal control problem to enable vehicle following with a safe constant time gap.
Also, a planning algorithm that searches over a discrete set of spline trajectories shall provide the initial trajectory and terminal manifold. 

\bibliographystyle{IEEEtran}
\bibliography{IEEEabrv, references}

\end{document}